\documentclass[sigplan, 10pt, screen, authorversion, nonacm]{acmart}

\AtBeginDocument{%
  \providecommand\BibTeX{{%
    \normalfont B\kern-0.5em{\scshape i\kern-0.25em b}\kern-0.8em\TeX}}}

\usepackage{graphicx}
\usepackage{subcaption}
\usepackage{ulem}
\usepackage{amsmath}
\usepackage{cases}
\usepackage{multirow}
\usepackage{colortbl}
\usepackage{booktabs}
\usepackage{hhline}
\usepackage{makecell} 
\usepackage{diagbox}
\usepackage{pifont}

\usepackage{hyperref}

\newcommand{\newtext}[1]{\textcolor{black}{#1}}

\makeatletter
\newif\if@restonecol
\makeatother

\usepackage[linesnumbered,ruled,vlined]{algorithm2e}
\usepackage{algpseudocode}
\usepackage{amsmath}

\SetKwInOut{KwIn}{Init}

\begin{document}
\raggedbottom

\title[]{\textsc{Klotski}: Efficient Mixture-of-Expert Inference via Expert-Aware Multi-Batch Pipeline}

\author{Zhiyuan Fang}
\affiliation{
  \institution{Sun Yat-sen University}
  \state{Zhuhai}
  \country{China}
}
\email{fangzhy27@mail2.sysu.edu.cn}

\author{Yuegui Huang}
\affiliation{
  \institution{Sun Yat-sen University}
  \state{Guangzhou}
  \country{China}
}
\email{huangyg35@mail3.sysu.edu.cn}

\author{Zicong Hong}
\affiliation{
  \institution{Hong Kong University of Science and Technology}
  \state{Hong Kong}
  \country{China}
}
\email{ziconghong@gmail.com}

\author{Yufeng Lyu}
\affiliation{
 \institution{Huawei Technologies Co. Ltd}
 \state{Shenzhen}
 \country{China}
}
\email{lvyufeng1@huawei.com}

\author{Wuhui Chen}
\affiliation{
  \institution{Sun Yat-sen University}
  \state{Zhuhai}
  \country{China}
}
\affiliation{
  \institution{Peng Cheng Laboratory}
  \state{Shenzhen}
  \country{China}
}
\email{chenwuh@mail.sysu.edu.cn}

\author{Yue Yu}
\authornotemark[1]
\affiliation{
  \institution{Peng Cheng Laboratory}
  \state{Shenzhen}
  \country{China}
}
\email{yuy@pcl.ac.cn}

\author{Fan Yu}
\affiliation{
  \institution{Huawei Technologies Co. Ltd}
  \state{Shenzhen}
  \country{China}
}
\email{fan.yu@huawei.com}

\author{Zibin Zheng}
\affiliation{
  \institution{Sun Yat-sen University}
  \state{Zhuhai}
  \country{China}
}
\email{zhzibin@mail.sysu.edu.cn}

\begin{abstract}
Mixture of Experts (MoE), with its distinctive sparse structure, enables the scaling of language models up to trillions of parameters without significantly increasing computational costs. However, the substantial parameter size presents a challenge for inference, as the expansion in GPU memory cannot keep pace with the growth in parameters. Although offloading techniques utilise memory from the CPU and disk and parallelise the I/O and computation for efficiency, the computation for each expert in MoE models is often less than the I/O, resulting in numerous bubbles in the pipeline.

Therefore, we propose \textsc{Klotski}, an efficient MoE inference engine that \newtext{significantly reduces pipeline bubbles }through a novel \textit{expert-aware multi-batch pipeline} paradigm. 
The proposed paradigm uses batch processing to extend the computation time of the current layer to overlap with the loading time of the next layer. 
Although this idea has been effectively applied to dense models, more batches may activate more experts in the MoE, leading to longer loading times and more bubbles. 
Thus, unlike traditional approaches, we balance computation and I/O time and minimise bubbles by orchestrating their inference orders based on their heterogeneous computation and I/O requirements and activation patterns under different batch numbers.
Moreover, to adapt to different hardware environments and models, we design a constraint-sensitive I/O-compute planner and a correlation-aware expert prefetcher for a schedule that minimises pipeline bubbles.
Experimental results demonstrate that \textsc{Klotski} achieves a superior throughput-latency trade-off compared to state-of-the-art techniques, with throughput improvements of up to 85.12$\times$.
\end{abstract}


\maketitle

\section{Introduction}

Owing to the rapid advancement of deep learning, large language models (LLMs) have demonstrated remarkable efficacy across various domains~\cite{brown2020language, devlin2018bert, zhang2022opt}. To facilitate model scalability without escalating the costs associated with training and inference, recent research has introduced sparsely activated Mixture-of-Experts (MoE) models~\cite{fedus2022switch, shazeer2017outrageously}.
MoE models typically replace the Feed-Forward Network (FFN) layers with MoE layers. For each input, only a subset of the parameters (i.e., experts) are sparsely activated for computation, rather than all parameters, significantly reducing computational cost. \newtext{Current research has demonstrated the superiority of the MoE architecture through extensive experiments~\cite{rajbhandari2022deepspeed, he2022fastermoe}.}

\newtext{
However, due to the skew between model parameter sizes and advances in hardware, MoE-based models, with their massive parameter counts, face more severe memory bottlenecks during inference than other LLMs. For example, DeepSeek-V2~\cite{deepseekai2024deepseekv2}, with 236 billion parameters, requires at least seven state-of-the-art (SOTA) GPUs (H100, with 80GB of memory each) for inference. 
Furthermore, the high cost of memory often makes it difficult to use such large models in more common environments such as personal computers and small servers, limiting the wider adoption of large models~\cite{xu2024survey, du2023sida, liu2024lightweight}.
This raises the question of how to deploy MoE models in resource-constrained environments where there is a significant gap between available GPU memory and model parameter sizes.}

\begin{figure}
  \centering
  \includegraphics[width=\linewidth]{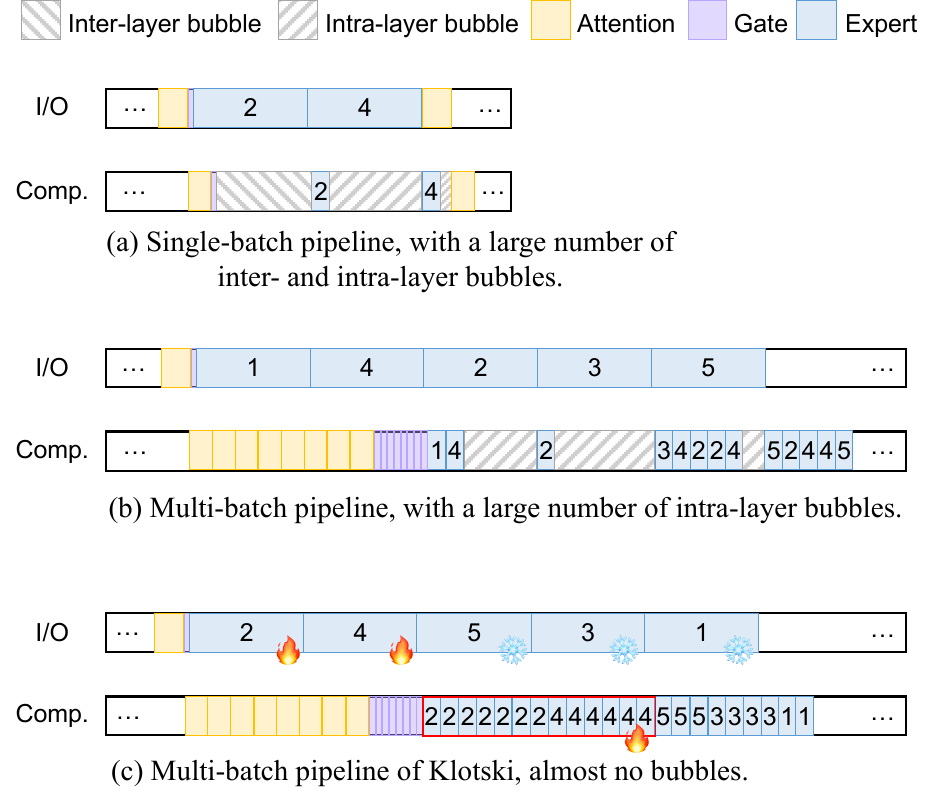}
  \caption{Comparison of three kinds of pipeline.
  We use multiple computations of the current layer to overlap the I/O of the next layer to reduce inter-layer bubbles and adjust the experts' computation order to reduce intra-layer bubbles.}
  \label{figure:1}
\end{figure}

Offloading is one of the current mainstream solutions for addressing memory optimization during the inference of LLMs~\cite{ren2021zero, sheng2023flexgen, eliseev2023fast, kamahori2024fiddler}. \newtext{It significantly reduces GPU memory requirements for LLM inference by offloading tensors not needed for the current computation.}
Applying offloading to MoE models is effective because the experts are sparsely activated, resulting in more parameters that can be offloaded during inference. 
Recent efforts\newtext{~\cite{xue2024moe, eliseev2023fast}} have proposed offloading strategies tailored for MoE models. \autoref{figure:1}(a) illustrates the basic paradigm of these methods: prefetching the next layer while computing the current layer to achieve partial overlap of I/O and computation. However, due to the sparse activation of experts, these methods often rely on the accuracy of expert prefetching. For instance, MoE-Infinity~\cite{xue2024moe} performs activation-aware expert prefetching and caching based on expert activation traces. SiDA~\cite{du2023sida} trains an offline expert predictor in a data-aware manner, achieving a prefetching accuracy of over 90\%.

However, significant \textit{inter-layer} and \textit{intra-layer bubbles} \newtext{(GPU stalls)} degrade performance due to the computation and I/O imbalance. 
Inter-layer bubbles occur because of the imbalance between the attention and expert layers. The large size of experts prevents the computation of the attention layer from sufficiently overlapping with the I/O of the expert layer. \newtext{In Mixtral-8$\times$7B~\cite{jiang2024mixtral}, using an NVIDIA 3090 to process a batch size of 16, the average attention computation is about 2.6 ms, while the single expert transmission time is about 21 ms.} Furthermore, when the number of experts selected by the gate exceeds one (as in Mixtral-8$\times$7B and DeepSeekMoE~\cite{dai2024deepseekmoe}, etc.), the I/O overhead for expert transmission multiplies, causing the GPU to wait more frequently. 
Intra-layer bubbles, on the other hand, result from an imbalance between computation and I/O within the expert layer. In the inference of dense models, the loaded FFN processes all sequences in the batch. However, in MoE models, each activated expert processes only a portion of the sequences in the batch but consume time to transfer multiple FFNs (each expert is an FFN). \newtext{For instance, processing a token with a single expert in Mixtral-8$\times$7B takes less than 1 ms, which is much less than the transmission delays.} This leads to substantial intra-layer bubbles between the computations of multiple experts.

Inspired by related work on dense models \cite{sheng2023flexgen, xuanlei2024hetegen}, a straightforward approach is to consider the computations of multiple batches simultaneously. This increases the total computation time, thereby allowing for the overlap of the I/O time for the next layer. Specifically, after loading the weights of a layer, they are shared across multiple batches, allowing consecutive computations within the current layer. This provides sufficient time for loading the weights of the subsequent layer, thereby significantly reducing inter-layer bubbles.

Despite this, considering the computations of multiple batches simultaneously also means increasing the diversity of the inputs to the MoE layer. Given the sensitivity of the gating mechanism to data variability\newtext{~\cite{xue2024openmoe, lin2024moe}}, the total number of activated experts may increase. As shown in \autoref{figure:1}(b), in addition to the experts activated in \autoref{figure:1}(a), experts 5 and 3 are also activated. Although multiple computations in the attention layer can overlap the I/O of some experts, more experts are activated, resulting in more intra-layer bubbles in the pipeline, due to the long I/O time for these experts.

To tackle this challenge, we propose an \textit{expert-aware multi-batch pipeline} paradigm. Specifically, based on current observations~\cite{xue2024openmoe, li2023accelerating}, there is a phenomenon in MoE inference where a few experts handle the majority of tokens, referred to as \textit{hot experts}. Correspondingly, other experts are termed \textit{cold experts}. Considering a large number of tokens across multiple batches, hot experts exhibit \newtext{high computational demand and low I/O demand}, while cold experts exhibit the opposite. By leveraging this complementary relationship, we can overlap the high I/O demand of cold experts with the high computational demand of hot experts, effectively minimizing intra-layer bubbles between experts. As illustrated in \autoref{figure:1}(c), we prefetch only the hot experts 2 and 4 and partition the computations of multiple batches by experts rather than by batches. Furthermore, we adjust the computation order of the experts, prioritizing the substantial computations of hot experts 2 and 4, providing more ample time for the transmission of cold experts 5, 3, and 1. This effectively compresses the intra-layer bubbles.

In this paper, based on the above paradigm, we propose \textsc{Klotski}, an MoE-oriented inference engine that can perform high-throughput inference in resource-constrained environments, achieving inference pipeline with near-zero bubble, as shown in \autoref{figure:1}(c). 
To summarize, we make the following contributions:

\begin{itemize}
\item We propose an expert-aware multi-batch pipeline paradigm that leverages the high computational demand and low I/O demand of hot experts to orchestrate multi-batch computations, aiming to minimize both inter-layer and intra-layer bubbles. 
\item We design a constraint-sensitive IO-compute planner to formulate execution plans for this paradigm in various environments.
\item We propose adaptive tensor placement and a correlation-aware expert prefetcher, enabling appropriate offloading and prefetching when dealing with different storage resources and MoE models. 
\item We implement the above strategies in \textsc{Klotski}, an MoE-oriented inference engine, which enables high-throughput inference of MoE with offloading.
\item To evaluate \textsc{Klotski}, we compare it with Hugging Face Accelerate~\cite{accelerate}, Deepspeed-FastGen~\cite{holmes2024deepspeed}, FlexGen~\cite{sheng2023flexgen}, \newtext{MoE-Infinity~\cite{xue2024moe}, and Fiddler~\cite{kamahori2024fiddler}}. The experimental results demonstrate that \textsc{Klotski} can make inference of MoE more efficiently, and achieve 85.12$\times$, 15.45$\times$, 2.23$\times$, \newtext{19.06$\times$, and 9.53$\times$} throughput improvement than that of the three aforementioned works, respectively.
\end{itemize}

\section{Background and Related Work}


\subsection{\newtext{MoE Architecture \& Inference}}
\label{section:2.1}

Since GShard \cite{lepikhin2020gshard} introduced MoE structure into Transformer models, its potential to enhance the performance of large language models has been evident. MoE has gradually become one of the mainstream structures of large language models. Prominent models like GPT-4 \cite{achiam2023gpt}, Gemini 1.5 \cite{team2023gemini}, and Mixtral-8$\times$7B all incorporate the MoE structure.

\begin{figure}[t]
  \centering
  \includegraphics[width=\linewidth]{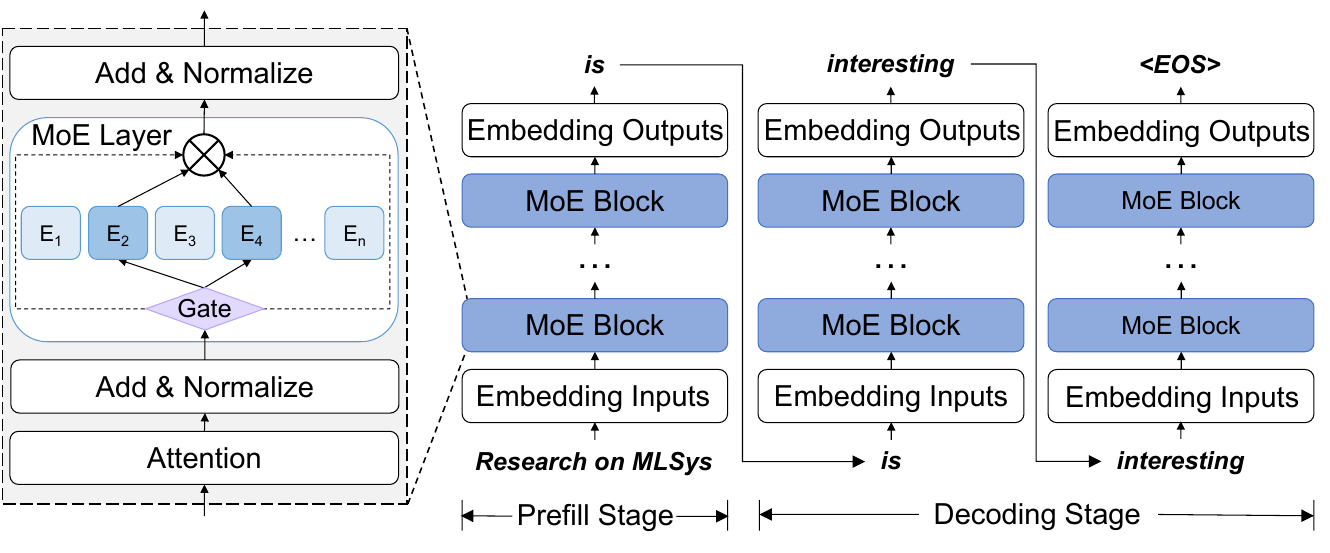}
  \caption{\newtext{Architecture and inference process of MoE models.}}
  \label{figure:2}
\end{figure}

\newtext{
The MoE architecture primarily consists of multiple MoE blocks, each containing an attention layer, an MoE layer, and two normalization layers, as illustrated in \autoref{figure:2}. }
The MoE layer comprises a gating network and multiple experts. 
The gating network is the key feature of the MoE architecture. It uses a softmax function to calculate the routing weights of each expert, then activates the top-k experts. Existing research~\cite{xue2024raphael, li2023accelerating, xue2024moe, lin2024moe} indicates that the expert activation path of each token can reveal its characteristics, \newtext{facilitating the prediction of future expert selections.
Each expert is an FFN, and experts are sparsely activated, making MoE a feasible approach to training larger models.} For each token, the final output of the MoE layer is a weighted sum of the outputs from the selected experts.

\newtext{
The MoE inference process, like other LLMs, follows an autoregressive approach~\cite{vaswani2017attention}, generating each new token based on the previous ones, as illustrated in \autoref{figure:2}. This process comprises two stages: prefill and decoding. In the prefill stage, the model processes the entire prompt simultaneously, which often leads to the activation of multiple experts. During the decoding stage, the model uses the previous token generated as input, iteratively generating new tokens until it generates the end-of-sequence (<EOS>) token or the maximum output length limit is reached.}

Some recent literature~\cite{zhai2023smartmoe, he2022fastermoe, rajbhandari2022deepspeed, li2023accelerating} has focused on the optimization of MoE. DeepSpeed-MoE~\cite{rajbhandari2022deepspeed} introduces a specialized MoE architecture called Pyramid-Residual MoE and employs staged knowledge distillation to obtain the Mixture-of-Students. This approach not only accelerates MoE training but also reduces inference latency and cost. Lina~\cite{li2023accelerating}, an extension of DeepSpeed-MoE, prioritizes all-to-all communication during training to enhance bandwidth and uses resource scheduling based on hot experts during inference to balance workload. However, these MoE systems primarily focus on latency-sensitive scenarios and place emphasis on MoE training. Klotski contributed to the memory optimization of MoE and is orthogonal to many of these works.

\subsection{Offloading in LLM Inference}
\label{section:2.2}
\begin{figure}[t]
  \centering
  \includegraphics[width=\linewidth]{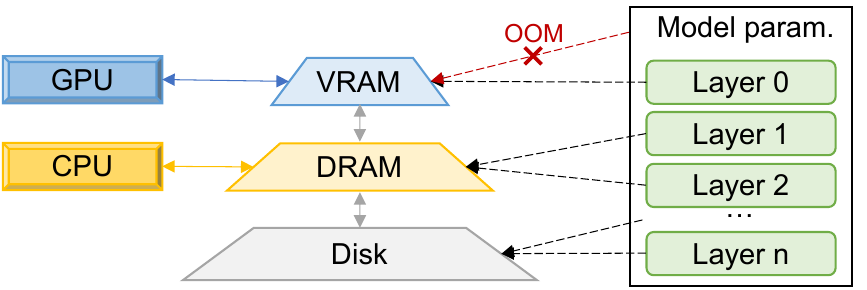}
  \caption{\newtext{Illustration of offloading an LLM in a multi-level storage system. Only a few layers of parameters can be placed in VRAM, and the rest are placed in DRAM and disk. param. refers to parameters.}}
  \label{figure:offloading}
\end{figure}

LLMs often have a large number of parameters, causing severe GPU memory bottlenecks during inference. Common memory optimisation techniques include quantization\newtext{~\cite{xiao2023smoothquant, kim2023mixture, frantar2022gptq}}, pruning\newtext{~\cite{ma2023llm, frantar2023sparsegpt}}, sparse attention\newtext{~\cite{xiao2023efficient, zhang2023h2o}}, etc. Among these, offloading is a particularly effective strategy in resource-constrained environments. \newtext{As shown in~\autoref{figure:offloading}, DRAM and disk often have at least dozens of times more memory than VRAM. When it is difficult to store all model parameters in VRAM (as in the red line), offloading strategies offload tensors not currently involved in computation to DRAM or disk, freeing up a significant amount of VRAM (as in the black lines). Consequently, offloading strategies allow LLM inference to be performed with extremely small memory footprints. However, because the I/O speed between VRAM and DRAM is slower than the GPU's computing speed, frequent I/O will cause large delays in inference.}

Early works~\cite{huang2020swapadvisor, peng2020capuchin} proposed leveraging swapping during the training of Deep Neural Networks (DNNs) to reduce GPU memory demands. ZeRO-Offload~\cite{ren2021zero} applied the offloading to the training of Transformer-based LLMs. ZeRO-Infinity~\cite{rajbhandari2021zero} extended this approach by incorporating disk as an additional offloading destination. DeepSpeed-Inference~\cite{aminabadi2022deepspeed}, which includes the ZeRO-Inference component, applies offloading techniques to the inference, enabling LLM inference in resource-constrained environments. FlexGen~\cite{sheng2023flexgen} significantly improves inference throughput by solving linear programming problems within the computational graph. HeteGen~\cite{xuanlei2024hetegen} leverages heterogeneous parallel computing between CPU and GPU, reducing the need for parameter I/O and achieving better resource allocation. STI~\cite{guo2023sti} maximizes IO/compute resource utilization through model sharding and elastic pipeline planning. 

However, most offloading systems mentioned above are designed for dense models and are thus inadequate for supporting MoE inference. Mixtral-offloading~\cite{eliseev2023fast} identified this limitation earlier and utilized LRU cache and quantization to quickly load a subset of experts, enabling the inference of Mixtral-8$\times$7B on consumer-grade hardware. Fiddler~\cite{kamahori2024fiddler} designed CPU-GPU orchestration specifically for MoE models, leveraging the computational power of CPUs to minimize data movement between the CPU and GPU. MoE-Infinity~\cite{xue2024moe} significantly reduced the latency overhead associated with offloading experts through activation-aware expert prefetching and caching mechanisms. Even if these works design accurate prefetch strategies or use the computing power of the CPU to accelerate the inference of MoE, it is still difficult to balance the gap between computation and I/O, resulting in lots of bubbles in the pipeline. In contrast, \textsc{Klotski} minimizes the bubbles in the pipeline by simultaneously arranging the computations of multiple batches and making full use of the computing resources of the GPU.

\section{Motivation}
\subsection{\newtext{Shortcomings of Existing Work}}
\label{section:3.2}
\newtext{
While MoE brings numerous advantages, it also faces significant challenges related to GPU memory usage due to the large number of experts. According to existing work, the percentage of experts' parameters in Switch Transformers can reach up to 99\%~\cite{du2023sida}. At the same time, these experts are sparsely activated. There is no need to keep them resident in expensive GPU memory. Therefore, the sparse activation feature of MoE makes offloading a highly suitable strategy to address its memory challenges. However, offloading is not a comprehensive solution, and will introduce new issues.}

Many existing approaches for MoE focus on improving the accuracy of expert prefetching\newtext{~\cite{du2023sida, kamahori2024fiddler, eliseev2023fast, hwang2024pre}}. However, due to the physical limitation that computation speed generally exceeds I/O speed\newtext{~\cite{guo2023sti, sheng2023flexgen}}, the I/O time for a single expert is significantly longer than the computation time. Thus, even with 100\% accurate prefetching, there would still be substantial pipeline bubbles due to the extended I/O time for experts.

\begin{table}[t]
\renewcommand{\arraystretch}{1.4}
\centering
\caption{\newtext{A comparison of the throughput (token/s) improvements when applying the I/O overlap strategy, designed for dense models, to a dense model (OPT) and an MoE model (Switch Transformers, decoder only). The compared results are shown in the same color block. The batch size is 4, and the sequence length is 512.}}
\resizebox{\linewidth}{!}{
\begin{tabular}{cccccc}
\Xhline{1pt}
\multicolumn{2}{c}{} & \multicolumn{2}{c}{\textbf{Dense model}} & \multicolumn{2}{c}{\textbf{MoE model}} \\ \cmidrule(lr){3-4}\cmidrule(lr){5-6}
\multicolumn{2}{c}{\multirow{-2}{*}{\textbf{Model}}} & \textbf{OPT-1.3B} & \textbf{OPT-6.7B} & \textbf{switch-base-16} & \textbf{switch-base-128} \\ \Xhline{1pt}
\multicolumn{2}{c}{Model Size} & 2.6 GB & 13.3 GB & about 2.2 GB & about 14 GB \\ \hline
 & Original & 14.3 & 3.3 & 13.63 & 2.52 \\ \cline{2-6} 
 & + Strategy & 43.09 & 12.15 & 28.79 & 7.31 \\ \hhline{~-----}
\multirow{-3}{*}{Thoughput} & Improvement & \cellcolor[HTML]{FCE4D6}201.33\% & \cellcolor[HTML]{E2EFDA}268.18\% & \cellcolor[HTML]{FCE4D6}111.23\% & \cellcolor[HTML]{E2EFDA}190.08\% \\ \Xhline{1pt}
\end{tabular}
}
\label{table:1}
\end{table}


In offloading strategies for dense models, efforts have been made to overlap I/O with computation\newtext{~\cite{sheng2023flexgen, xuanlei2024hetegen, guo2023sti}}. One effective method for achieving high-throughput inference is to overlap the I/O of the next layer with multiple computations of the current layer, keeping the GPU almost always in the computation state~\cite{sheng2023flexgen}. \newtext{We applied this method to the inference of a dense model (OPT) and an MoE model (Switch Transformers) of similar model size, with results shown in \autoref{table:1}. 
The results show that the improvement of using this strategy for dense models is significantly higher than for MoE models. This is because it uniformly prefetches the next layer during the computations of the current layer, without considering the special I/O resource demands of the MoE layer, which contains multiple FFNs.} Other strategies designed for dense models are the same.
Thus, direct application of the existing SOTA method, designed for dense models, to MoE models often results in a loss of performance. 

From the above, we know that existing offloading strategies are insufficient for MoE models. There is still a need for an efficient offloading strategy that can extremely compress the bubbles in the pipeline for MoE inference. 




\subsection{\newtext{A Strawman Offloading Strategy for MoE Models}}
\label{section:3.3}
\begin{figure}[t]
  \centering
  \includegraphics[width=\linewidth]{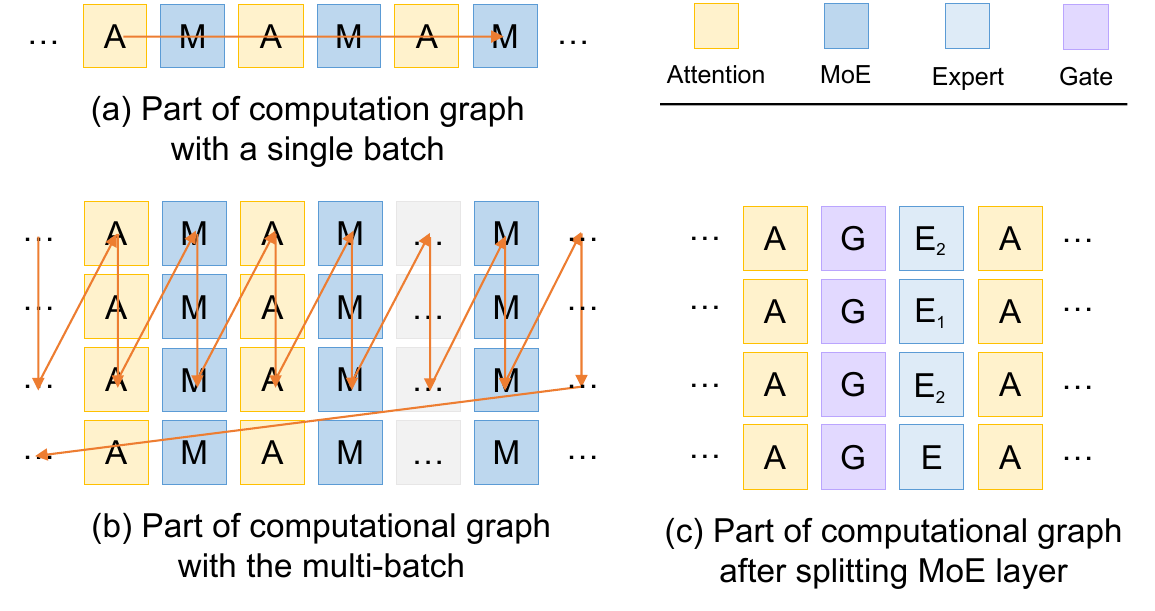}
  \caption{\newtext{Construction process of strawman offloading strategy designed for MoE Models. Each row represents a batch.}}
  \label{figure:strawman}
\end{figure}

\newtext{
To better adapt the aforementioned I/O overlapping strategy for MoE inference, we propose a strawman offloading strategy, the outline of whose construction process is shown in~\autoref{figure:strawman}. Two normalization layers are incorporated into the attention and MoE layers, respectively. Then we explain it incrementally from a simple offloading strategy as follows.}

\newtext{
Firstly, as shown in~\autoref{figure:strawman}(a), a simple offloading strategy is executing computations sequentially following the architecture of the MoE model, prefetching parameters of the next layer while computing the current layer. However, due to the slower I/O speed compared to computation speed, eliminating bubbles in the pipeline of a single batch is unfeasible, leading to low inference efficiency.}

\newtext{
Consequently, inspired by FlexGen~\cite{sheng2023flexgen}, we opted to expand the computational graph to multiple batches, as illustrated in~\autoref{figure:strawman}(b). After loading the weights of a certain layer, the strawman strategy executes computations of multiple batches while loading the weights of the next layer in parallel, thereby achieving more overlap. Nevertheless, it also presents challenges: the MoE layer, consisting of a gate and several experts, is large, resulting in a long wait for the I/O of the entire layer. Moreover, not all experts are involved in the computation, resulting in unnecessary I/O.}

\begin{figure*}
  \centering
  \includegraphics[width=\linewidth]{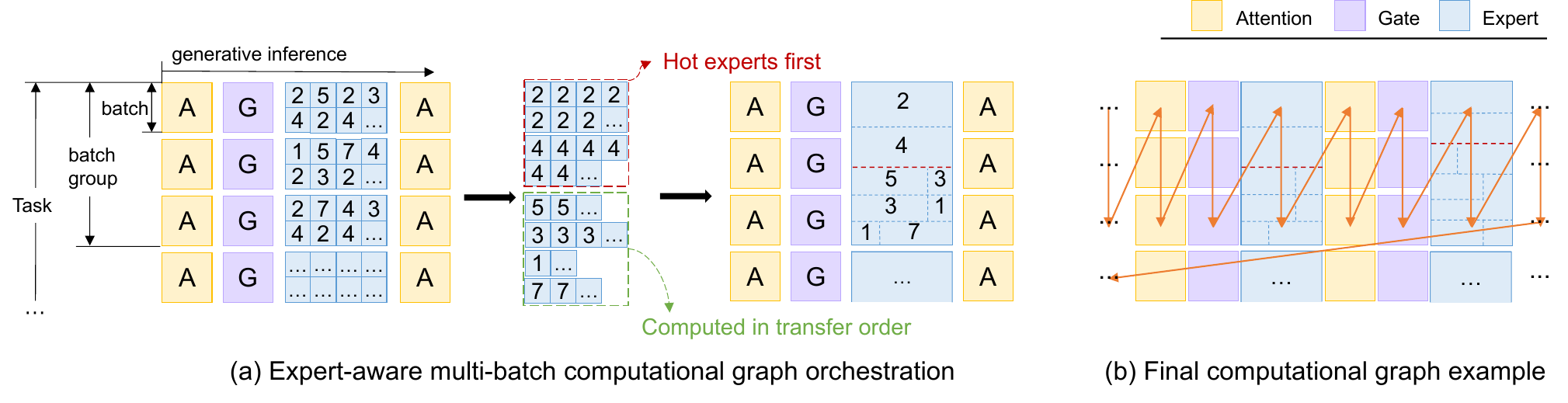}
  \caption{Expert-aware multi-batch computational graph.}
  \Description{Expert-aware multi-batch computational graph.}
  \label{figure:6}
\end{figure*}

\newtext{
Furthermore, to solve the problem, we partitioned the MoE layer into a gate layer and an expert layer, as depicted in~\autoref{figure:strawman}(c). During the computations of the attention layer, the strawman strategy only prefetches the weights of the gate and a subset of experts, effectively reducing inter-layer bubbles. Then, after the computations of each gate, we check whether the selected expert has already been selected. If not, we initiate the transfer of the expert. However, we still face two problems: (1) how to determine which and how many experts to prefetch, (2) as illustrated in Figure \ref{figure:6}(c), assuming $E_2$ has already been prefetched during the computations of the attention layer, while $E_1$ is still undergoing transfer. The order of computations ($E_2 \rightarrow E_1 \rightarrow E_2$) may stall in the second step, while $E_2$ in the third step could have been computed directly, resulting in unnecessary intra-layer bubbles.}

\begin{figure}[t]
  \centering
  \includegraphics[width=\linewidth]{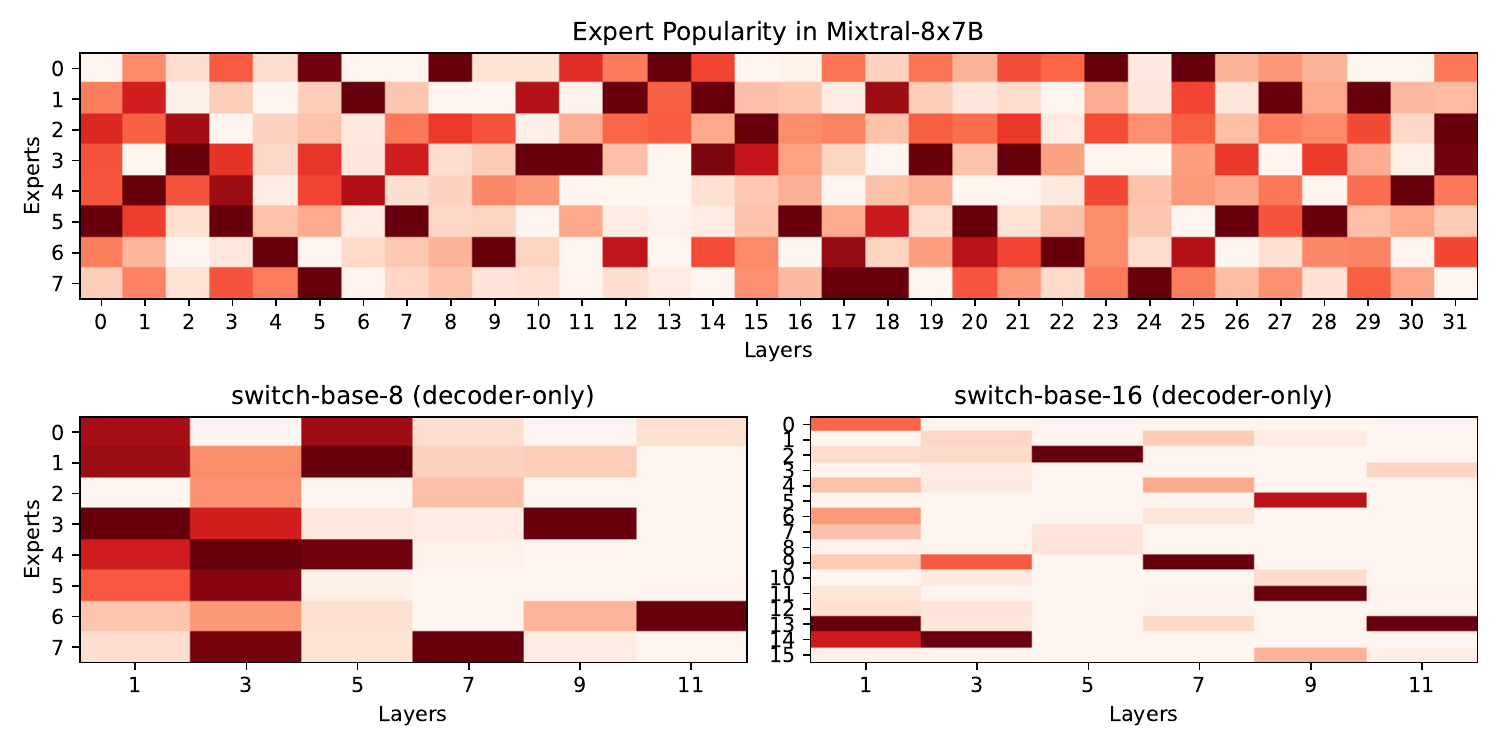}
  \caption{The expert heatmaps in Mixtral-8$\times$7B, \newtext{decoder part of switch-base-8 and switch-base-16. The darker the color, the higher the frequency of selection.}}
  \label{figure:heatmaps}
\end{figure}

\newtext{
For the problem (1), during MoE inference, there is a phenomenon of hot experts, where a few experts handle the majority of tokens~\cite{li2023accelerating, lin2024moe}.} As shown in \autoref{figure:heatmaps}, we recorded the expert selections for Switch Transformers and Mixtral-8$\times$7B. It is evident that, with a high probability, \newtext{tokens will be routed to hot experts. Furthermore, K (K equals k in top-k) experts usually cover most of the inputs. For example, during inference with Mixtral-8$\times$7B, which uses the top-2 gate, tokens tend to select experts 1 and 3 in layer 14, with a total ratio of 53.7\%. Similar situations can be clearly observed in other layers. Therefore, while performing multiple computations of the attention layer, we prefetch the gate and K experts as hot to reduce inter-layer bubbles.} 

\newtext{
For the problem (2), the strawman strategy adjusts the computation order of experts,} overlaps the I/O of cold experts with the computations of hot experts. Since hot experts handle the majority of tokens across multiple batches, their computation time can provide more time for the transfer of subsequent experts.

\newtext{
\textbf{Challenges.} While the strawman strategy provides a comprehensive offloading approach tailored for MoE models, we encountered several challenges in its practical application. First, experts are data-sensitive\newtext{~\cite{zhai2023smartmoe}}, meaning that the hot experts may change when the input tokens vary; thus, it is challenging to dynamically identify hot experts. Second, in the actual inference process, the number of experts involved in computations is significantly higher than depicted in~\autoref{figure:strawman}(c); thus, it is challenging to orchestrate multi-batch expert computations. Third, the hardware environment for model inference is very diverse; thus, it is challenging to provide efficient inference in uncertain hardware environments.}


\section{Overview}

\begin{figure}
  \centering
  \includegraphics[width=\linewidth]{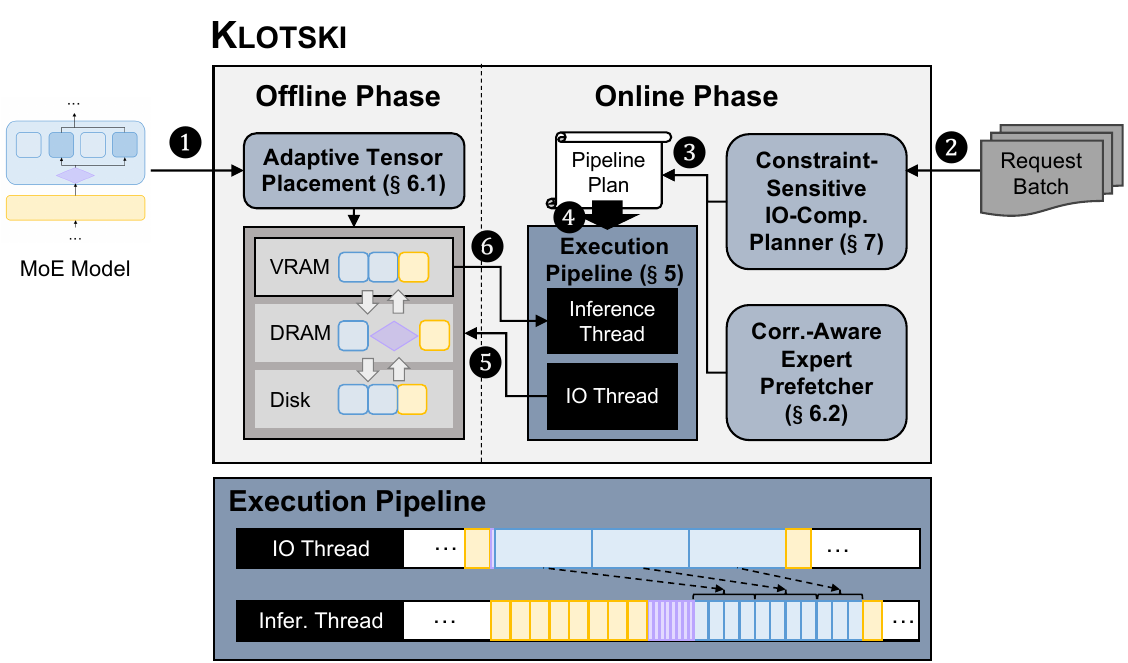}
  \caption{System overview of \textsc{Klotski}. Comp. refers to Compute, Corr. refers to Correlation, and Infer. refers to Inference. The blue, purple, and yellow graphics represent the expert, gate, and attention layers, respectively. The dotted line in the pipeline indicates that the computations within the curly brackets belong to the expert.}
  \Description{System overview of \textsc{Klotski}. Comp. refers to Compute, Corr. refers to Correlation, and Infer. refers to Inference. The blue, purple, and yellow graphics represent the expert, gate, and attention layers, respectively.}
  \label{figure:overview}
\end{figure}

To solve the above challenges, we propose \textsc{Klotski}, an inference engine designed for MoE that enables high-throughput MoE inference in resource-constrained environments. We show the system overview of \textsc{Klotski} in \autoref{figure:overview}.

Firstly, during the offline phase, we aggregate heterogeneous memory from GPU, CPU, and disk for model deployment. We adaptively sense the memory limits in the current environment and allocate the MoE model tensors across the heterogeneous memory \newtext{(\raisebox{-0.15ex}{\ding{182}})} accordingly. In the online phase, when request batches are inputted \newtext{(\raisebox{-0.15ex}{\ding{183}})}, the constraint-sensitive I/O-compute planner formulates a pipeline plan based on the current hardware constraints \newtext{(\raisebox{-0.15ex}{\ding{184}})}. If the MoE model is performed for the first time, the correlation-aware expert prefetcher generates an expert correlation table during the warm-up process to guide expert prefetching. 

According to the settings in the pipeline plan, \textsc{Klotski} executes computations following the expert-aware multi-batch pipeline paradigm to achieve an execution pipeline with minimal bubbles \newtext{(\raisebox{-0.15ex}{\ding{185}})}. During inference, the I/O thread dynamically manages the transfer of tensors across heterogeneous memory \newtext{(\raisebox{-0.15ex}{\ding{186}})}. The inference thread reads the corresponding tensors from VRAM for computations \newtext{(\raisebox{-0.15ex}{\ding{187}})}. Additionally, the inference thread continuously updates the expert correlation table to further capture the data tendencies of the task.

\section{Expert-aware Multi-batch Pipeline Paradigm}
\label{section:5}
We aim to develop a pipeline that minimizes all bubbles to maximize GPU utilization. To achieve this, we propose an expert-aware multi-batch pipeline paradigm, \newtext{which is designed based on zig-zag block schedule~\cite{sheng2023flexgen}}. By considering the computations of multiple batches simultaneously, this paradigm enables weight sharing and orchestrates the multi-batch computational graph around the experts to reduce bubbles. A partial computational graph is illustrated in \autoref{figure:6}, where each row corresponds to the computations of one batch, and the multiple batches are considered together as a batch group. Ultimately, this results in a nearly bubble-free pipeline, as shown in \autoref{figure:7}. In the following, we will detail this paradigm from two perspectives: minimizing inter-layer bubbles and minimizing intra-layer bubbles.

First, minimizing inter-layer bubbles. Inter-layer bubbles primarily occur between the attention layer and the MoE layer. During the computations of multiple batches in the attention layer, \textsc{Klotski} prefetches only the weights of the gate and the hot experts, rather than the entire MoE layer. Because overlapping the I/O for the entire MoE layer is challenging, and \autoref{formula:1} must be satisfied.
\begin{equation}
\label{formula:1}
  n * t_{c\_A} \geq t_{I/O\_MoE}
\end{equation}
Here, \( n \) represents the number of batches in a batch group, \( t_{c\_A} \) denotes the computation time of an attention layer for a batch, and \( t_{I/O\_MoE} \) is the time required to transfer the entire MoE layer. \autoref{formula:1} clearly necessitates a large \( n \) to hold true, which would introduce a significant amount of KV cache. \newtext{What's more, due to the nature of sparse activation, some experts may not be activated, even when multiple batches are being processed at the same time.} Loading them all into VRAM not only wastes resources but also increases latency. In contrast, only overlapping the I/O for the gate and hot experts is easier and more effective, which just needs to satisfy \autoref{formula:2}. 
\begin{equation}
\label{formula:2}
  n * t_{c\_A} \geq t_{I/O\_G} + K * t_{I/O\_E}
\end{equation}
Here, \( t_{I/O\_G} \) and \( t_{I/O\_E} \) represent the transfer times for the gate and a single expert, respectively. \( K \) equals \( k \), the number of experts selected by the top-k gate, usually 1 or 2. Hot experts are chosen because they are likely engaged in most of the computations (see \autoref{figure:heatmaps}), which provides an opportunity to minimize intra-layer bubbles subsequently. Additionally, during the computations of the gate, no prefetching is done. Instead, it is determined whether each gate-selected expert is a hot expert or one that has already been transferred. If not, the transfer of that expert is initiated immediately.

\begin{algorithm}[t]
  \caption{Schedule Algorithm of the Paradigm.}
  \label{algorithm:1}
  \KwIn{Generate length $l$, number of layers $n\_layer$, number of batches $n\_batch$, hidden state $h$, KV cache $c$. \newtext{The indices $i,j,k$ indicate that the it is processing the $i$-th token, performing computations at the $j$-th layer for the $k$-th batch.}}
  \For{$i < l$}{
    \For{$j < n\_layer$}{
        \If{$layers[j]\ is\ not\ Gate$}
        {
            $load(layers[j+1])$
        }
        \If{$layers[j]\ is\ Expert\_Layer$}
        {
            $load(c[i][j+1][0])$ \\
            \begin{tabular}{@{}l}
            \Comment{Experts process all tokens across batches.} \\
            $compute(layers[j])$ \\
            \end{tabular} \\
            $store(h[i][j])$ \\
            $load(h[i][j+1][0])$
        }
        \Else{
            \Comment{Non-expert process each batch vertically.} \\
            \For{$k < n\_batch$}
            {
                $sync(load\_cache\_stream)$ \\
                $load(h[i][j][k+1],\ c[i][j][k+1])$ \\
                $compute(layers[j][k])$ \\
                $sync(store\_cache\_stream)$ \\
                $store(h[i][j][k],\ c[i][j][k])$
            }
        }
        $sync(load\_weight\_stream)$
    }
  }
\end{algorithm}

Second, minimizing intra-layer bubbles. As illustrated in the left panel of \autoref{figure:6}(a), the sequence of experts shows that hot experts 2 and 4 have already been prefetched, while experts 5 and 3 are still undergoing transfer. Thus, the sequence of computations [2523424…] would result in the GPU stalling at positions 5 and 3, due to the incomplete transfer of data at these locations. However, computations involving experts 2 and 4 could proceed immediately. To reduce such unnecessary delays, we further adjust the order of expert computations across multiple batches, allowing computations involving the same experts to run continuously and prioritizing computations of hot experts. Since hot experts are transferred to GPU memory first and engaged in more computations, this adjustment allows more time for the transfer of experts still being loaded. After the computations for hot experts, the remaining experts compute in the order they are transferred. Additionally, experts that have completed all computations are offloaded immediately, rather than waiting for the entire layer's computations to finish, to reduce peak GPU memory usage.

Finally, \textsc{Klotski} executes computations according to the computational graph shown in \autoref{figure:6}(b), sharing the loaded weights across multiple batches. This approach not only reduces the number of I/O operations to approximately $1/n$ of the original but also overlaps the time for each I/O, resulting in an almost bubble-free pipeline as illustrated in \autoref{figure:7} and significantly improving throughput. The algorithm details of this paradigm are formulated in \autoref{algorithm:1}. First, since hot experts are already prefetched during the attention layer, we do not perform prefetching in the gate layer (line 3), instead, the real-time transfer of experts is based on its results. Second, experts process all tokens across batches (line 5), since the computations of the expert layer are divided by experts rather than by batches. Third, the non-expert layer processes each batch sequentially (line 11), prefetching the necessary activations, key-value caches, etc., for the corresponding batch. Additionally, we synchronize the transfers of various streams using the $sync()$ function.

\section{Tensor Management}
\subsection{Adaptive Tensor Placement}
\label{subsection:6.1}


\textsc{Klotski} constructs a multi-level heterogeneous memory space consisting of VRAM, DRAM, and disk to meet the storage demands of MoE models in resource-constrained environments. Then, we propose an adaptive tensor placement, which intelligently allocates tensors based on the available memory resources in the current environment, thereby enhancing the utilization of existing resources.

Firstly, the GPU memory is primarily used to store necessary tensors required for current computations and prefetched tensors. When there is ample free GPU memory available, it can be further utilized to reduce some I/O operations. Specifically, we can choose storage locations for different types of tensors such as expert, gate, attention, KV cache, and activation. Furthermore, support is provided for layer granularity distribution. For example, placing the experts of the first three layers in VRAM, the experts of the next twenty layers in DRAM, and the remaining in disk.

Secondly, inactive tensors can be offloaded to either CPU memory or disk. We prioritize allocating CPU memory to experts. This is because the MoE layer faces the challenge that the experts requested by the gating function cannot be accurately predicted in advance. Therefore, when handling tasks with large batch sizes, it is highly likely that immediate transfers of experts will be needed, necessitating the rapid transfer of the required expert to GPU memory. Considering the faster transfer bandwidth of CPU memory, which provides quicker response times, we prioritize placing expert parts in CPU memory.

Additionally, when sufficient CPU memory is available, we use $pin\_memory$ to achieve faster CPU-GPU communication. When CPU memory is insufficient and disk usage is necessary, to reduce the GPU getting tensors from disk, which is slow, we dynamically maintain tensors in the CPU memory. Specifically, we dynamically manage tensors for a fixed number of layers \( L \) within the limited CPU memory. As the computation proceeds to layer \( i \), the GPU prefetches tensors for layer \( i+1 \) from CPU memory, while the CPU prefetches tensors for layer \( i+L \) from the disk and removes tensors for layer \( i \). This strategy effectively utilizes the idle CPU-disk bandwidth, thereby reducing the interaction between GPU and disk.

\subsection{Correlation-aware Expert Prefetcher}
\label{section:6.2}
For dense models, the offloading strategies can directly prefetch the next layer. However, it is different for MoE models. Only after completing the computation of the gate can the activated experts be determined, making it challenging to design a unified prefetching strategy.

\begin{figure}
  \centering
  \includegraphics[width=\linewidth]{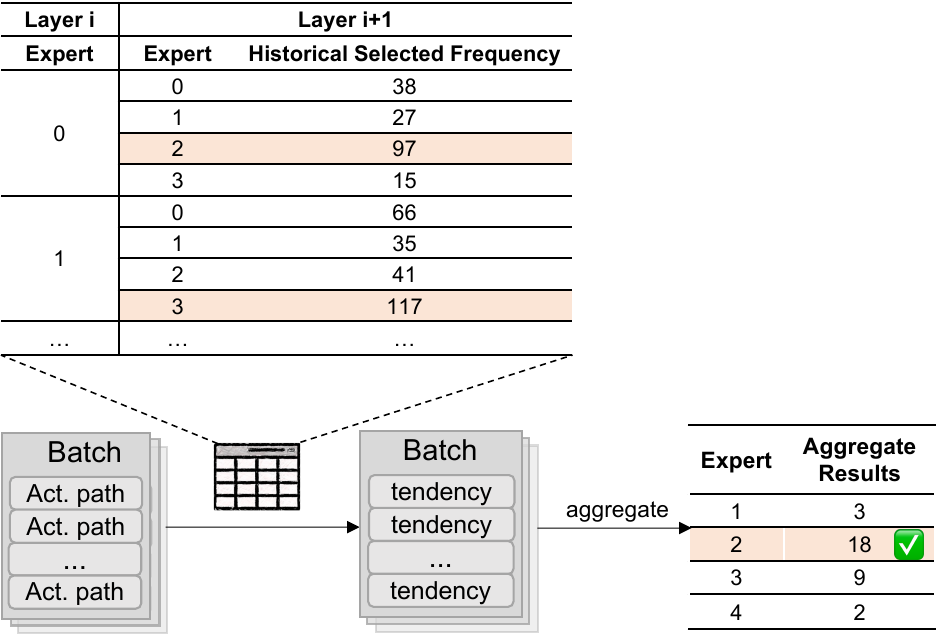}
  \caption{An example of the expert correlation table. Each expert layer has four experts. The gate selects the top-1 expert. The correlation path length $l$ is 1.}
  \label{figure:ECtable}
\end{figure}

To address this, \textsc{Klotski} design a correlation-aware expert prefetcher. In \autoref{section:5}, the prefetched experts need to engage in most computations across multiple batches to reduce intra-layer bubbles effectively. As illustrated in \autoref{figure:heatmaps}, there are hot experts in the inference of MoE, where a few experts cover the majority of computations. Therefore, the prefetching targets for the MoE layer are the gate and hot experts. 
Since MoE is data-sensitive and hot experts may vary with different inputs, we establish a data-aware expert correlation table to identify the hot experts that tokens in the current multi-batch tend to select. Specifically, we record the correlations (i.e., frequency relationships) between experts activated by tokens at different layers through pre-run, resulting in a table. During inference, we use this table to determine each token's expert tendency in the current layer based on its selections in the previous \( l \) layers. The larger the value of \( l \), the more accurate the prefetching. This process is illustrated in \autoref{figure:ECtable}, where each layer has four experts, the gate selects the top-1 expert, and \( l = 1 \). For the expert activation path of each token in the multi-batch, we look up the table to determine their expert tendencies in the current layer. We then aggregate the tendencies of all tokens across multiple batches and select the top-K experts for prefetching. \newtext{K is by default equal to k in top-k because, based on the observation in~\autoref{section:3.3}, K experts will generally cover the majority of the token computations.}

In addition, the expert correlation table is updated during the inference so that expert prefetching can become more and more accurate, as the table is continuously updated to understand the tasks at hand. To prevent the prefetching tendencies of other tasks from influencing current tasks, we refrain from saving the updates to the file.

On the other hand, for non-expert tensors, we adopt a prefetching strategy similar to that used for dense models, where we prefetch the tensor during the computations of the previous layer. This is because non-expert tensors are involved in computation only once during a forward pass and remain inactive at other times.

\section{Constraint-Sensitive I/O-Compute Planner}
\label{subsection:6.2}

\begin{figure}
  \centering
  \includegraphics[width=\linewidth]{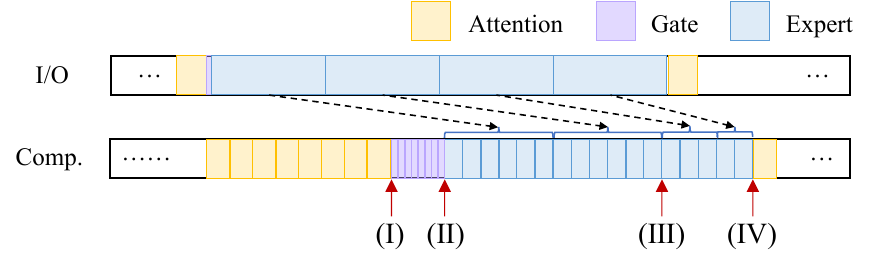}
  \caption{\newtext{Multi-batch} pipeline of \textsc{Klotski}.}
  \label{figure:7}
\end{figure}

\textbf{Planning Goal:} To minimize the total time \( T \) required to complete tasks under existing resource constraints, achieving an almost bubble-free pipeline as illustrated in \autoref{figure:7}.

\begin{equation}
\label{formula:min}
\begin{aligned}
    {min} \quad & {T = T_c + T_b} \\
    {s.t.} \quad & {M_{usage} < M_{GPU} + M_{CPU} + M_{disk},} \\
                      & {M_{peak\_GPU} < M_{GPU}}
\end{aligned}
\end{equation}

The total time \( T \) is primarily composed of two parts: \( T_c \) and \( T_b \), representing the total computation time and the total time occupied by bubbles, respectively. \( T_c \) mainly depends on hardware conditions. Our objective is to minimize \( T_b \) under the constraints of available memory, making it approach zero, \newtext{as shown in \autoref{formula:min}}. In our system, the reduction of \( T_b \) is primarily influenced by two factors: (1) the placement of the tensors and (2) the batch size and the number of batches included in the batch group, denoted as \( n \). Effective model placement can maximize the utilization of existing storage resources, thereby reducing some of the I/O demands, as considered in \autoref{subsection:6.1}. The batch size is typically a multiple of 4, leaving limited options for selection. However, determining the value of \( n \) is crucial. If \( n \) is too large, it will introduce a significant KV cache. Conversely, if \( n \) is too small, the total computation time for \( n \) batches may not overlap effectively with the I/O time of the next layer.

To investigate the value of $n$, our primary focus lies on the inter-layer overlap and intra-layer overlap in each MoE block. In \autoref{figure:7}, we have inserted several arrows indicating the points where a specific tensor needs to start computing. These are interpreted as follows: 
(\uppercase\expandafter{\romannumeral1}) indicates the point where gate computations will begin, 
(\uppercase\expandafter{\romannumeral2}) marks the start of the computations for hot experts, 
(\uppercase\expandafter{\romannumeral3}) signifies the beginning of the computations for cold experts, and 
(\uppercase\expandafter{\romannumeral4}) denotes the initiation of the next attention layer's computations. 
These arrows collectively suggest that the corresponding tensor transmission must be completed before these points to ensure that I/O and computation are fully overlapped. We list the four key positions with the respective inequalities that must be satisfied as follows.
\begin{numcases}{}
\label{Inequality Group}
n * t_{c\_A} \geq t_{I/O\_G}\\
n * (t_{c\_A}+t_{c\_G}) \geq t_{I/O\_G}+K*t_{I/O\_E}  \\
n * (t_{c\_A}+t_{c\_G})+t_{c\_hot-E} \geq t_{I/O\_G}+(K+1)t_{I/O\_E} \\
\notag n * (t_{c\_A}+t_{c\_G})+t_{c\_hot-E}+ \sum_{i\in Q}^{Q} t_{c\_E_i} \geq \\
\qquad \qquad \qquad t_{I/O\_G}+(K+len(Q))t_{I/O\_E}+t_{I/O\_A}
\end{numcases}
where K denotes the number of prefetched hot experts, $t_{c\_A}$, $t_{c\_G}$, $t_{c\_topk-E}$, $t_{c\_E_i}$, denote the time to compute attention, gate, hot experts and expert $i$, respectively, $t_{I/O\_A}$, $t_{I/O\_G} $, $t_{I/O\_E}$, denote the time to transfer attention weights, gate weights and weights of a single expert, respectively. The I/O times and computation times vary with hardware, model, and batch size. Additionally, the length of the queue \(Q\) of activated experts per layer is not fixed. \newtext{We determine the length of each layer of Q based on statistical data.}

In response to this, our planner operates primarily in two stages: (1) Measurement of the current hardware capability. Before the inference with an MoE model, \textsc{Klotski} measures the computation times and transmission durations of the model's various layers based on their shapes, data types, and other relevant information in the current environment. These results are cached locally. (2) Constraint solving. \textsc{Klotski} applies the measured data to the constraints from the inequality group to determine the optimal value of \( n \). Assuming the final result is \( n \geq x \), then \( n = \lceil x \rceil \). At this point, \( n \) ensures a pipeline without bubbles. Further increasing \( n \) might improve throughput, but the increase will be marginal because the pipeline is already near bubble-free. However, this would introduce a significant burden of massive KV caches on storage. Therefore, \( n \) should be set to the smallest integer that satisfies the inequality group. Additionally, if \( n \) becomes excessively large, manual adjustments to the strategy may be necessary. Since \( n \) is a positive integer, this process is not challenging.

Subsequently, we examine the potential outcomes of this strategy, considering both the most favorable and the least favorable scenarios. In the optimal scenario, all tokens select hot experts, thereby eliminating the need to consider inequalities (4) and (5). On the other hand, the worst-case scenario emerges when all tokens select cold experts, encompassing all other experts. In such instances, the value of $t_{c\_hot-E}$ is equal to zero, rendering the prefetching strategy ineffective. Inadequate $n$ may lead to a few intra-layer bubbles. However, intuitively, the probability of encountering such a worst-case scenario is very low.

\textbf{Compression} 
In particular, quantization and sparse attention are particularly well-suited for our work because they not only further reduce memory requirements but also decrease the amount of data transferred between heterogeneous memory, aiding in bubble reduction. \newtext{Therefore, we incorporated two effective methods as options.}

\textbf{(1) Quantization} Existing knowledge indicates that the sparsely activated expert weights are highly robust to quantization~\cite{kim2023mixture}. They can be quantized to 3 bits without additional training or calibration data. Since the majority of weights in MoE models belong to experts, quantizing the experts can significantly reduce memory requirements and I/O delays with minimal precision loss. Before computation, we convert the tensors back to their original precision through dequantization, further mitigating precision loss.

More specifically, we employ Half-Quadratic Quantization (HQQ)~\cite{badri2023hqq}. Quantization and dequantization are primarily achieved using the \autoref{equation:8}. 
\begin{equation}
\label{equation:8}
  Q_{z,s}(W) = W_q = round(W / s + z), \ Q^{-1}_{z,s}(W_q) = s(W_q - z)
\end{equation}
among them, the zero point $z$ and the scale $s$ are quantization parameters, which are determined through a robust optimization formula like \autoref{equation:10}.
\begin{equation}
\label{equation:10}
  \underset{z, \mathrm{~s}}{\operatorname{argmin}}\left(W-Q_{z, s}^{-1}\left(Q_{z, s}(W)\right)\right.
\end{equation}

In our study, to strike a balance between accuracy and transmission speed, we opt to preset that quantize both expert and attention tensors to 4 bits, using a group size of 64 and a zero scale group size of 128.

\textbf{(2) Sparse Attention} In this work, concurrently processing multiple batches necessitates storing a large amount of KV cache. Sparse attention not only reduces the KV cache size but also decreases the cost of transferring it across heterogeneous memory. We incorporate the attention mechanism from StreamingLLM~\cite{xiao2023efficient}, which focuses only on the initial sink tokens and neighboring tokens to achieve effective inference. This approach also enhances the model's ability to handle long sequences efficiently. \newtext{Additionally, this is optional as there are many models that have sparse strategies natively.}

\section{Implementation}

We implement \textsc{Klotski} on top of PyTorch~\cite{paszke2019pytorch} and Hugging Face Transformers~\cite{wolf2019huggingface} with over 3k LOC of Python.
Expert-aware multi-batch pipeline paradigm is implemented on top of FlexGen~\cite{sheng2023flexgen}.

\textbf{Expert Correlation Table.} 
We acquire input data by randomly sampling from wikitext-2~\cite{merity2016pointer}. Subsequently, we conduct inference with a batch size of 8 and a sequence length of 512. Expert selections during the inference are recorded and tabulated in JSON format. The choice of small batches is deliberate to avoid excessively large statistical values, which would render updates to the expert correlation table meaningless. We set the activation path length \( l = 1 \) because we do not heavily rely on the accuracy of expert prefetching. A larger number of batches in a batch group already allows us to overlap communication and computation. Increasing \( l \) would add dimension to path recording, which increases the complexity of the table lookup and memory occupation.

\textbf{Overlapping Computation and I/O.} 
Klotski achieves I/O-computation overlap by orchestrating four CUDA streams: one for prefetching weights, another for transferring expert weights based on gating network results, a third for prefetching KV cache, and the last for storing new KV cache. Each stream operates asynchronously, executing its designated task independently. When certain data is needed, the corresponding stream will be synchronized.

\begin{figure*}
  \centering
  \includegraphics[width=0.95\linewidth]{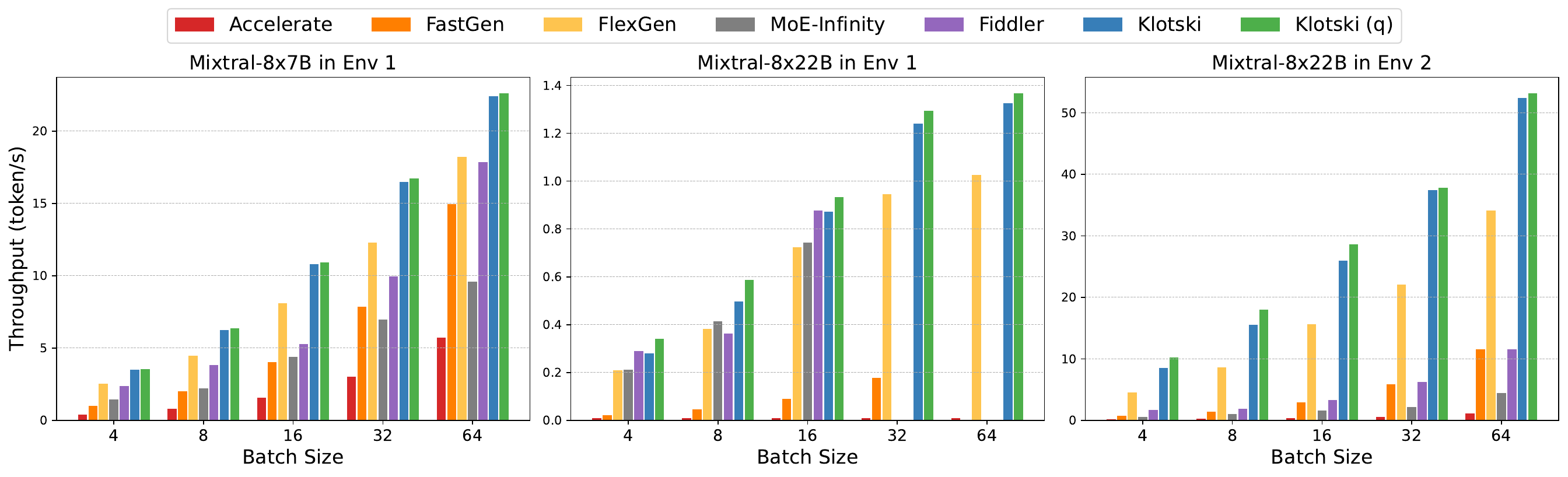}
  \caption{\newtext{Throughput comparison between \textsc{Klotski} and baselines in different scenarios. (q) means that quantization and dequantization are used.}}
  \label{figure:9}
\end{figure*}

\section{Evaluation}
\subsection{Experimental Setup}

\textbf{Hardware.} 
We evaluate \textsc{Klotski} in two different environments, as shown in the \autoref{table:4}. We don't care about the speed of disk reading in environment 2, \newtext{because} there is enough CPU memory.

\begin{table}[h]
\renewcommand{\arraystretch}{1.4}
\caption{Hardware environments for evaluation.}
\centering
\resizebox{\linewidth}{!}{
\begin{tabular}{c|cc|cc}
\Xhline{1pt}
\multirow{2}{*}{\textbf{Hardware}} & \multicolumn{2}{c|}{\textbf{Environment 1}} & \multicolumn{2}{c}{\textbf{Environment 2}} \\ \hhline{~----}
                                   & \textbf{Model}          & \textbf{Memory}  & \textbf{Model}           & \textbf{Memory} \\ \Xhline{1pt}
\textbf{GPU}                       & NVIDIA RTX 3090         & 24 GB            & NVIDIA H800              & 80 GB           \\ \hline
\textbf{CPU}                       & Intel Xeon Gold 5318Y   & 256 GB           & Intel Xeon Platinum 8470 & 800GB     \\ \hline
\textbf{Disk}                      & SSD                     & 2T               & SSD                      & 1T              \\ \hline
\textbf{PCIe}                      & \multicolumn{2}{c|}{4.0 x 16}               & \multicolumn{2}{c}{5.0 x 16}               \\ \hline
\textbf{Disk Read}                 & \multicolumn{2}{c|}{1 GB/s}                 & \multicolumn{2}{c}{/}                 \\ \Xhline{1pt}
\end{tabular}
}
\label{table:4}
\end{table}

\textbf{Models and Datasets.} 
We evaluate \textsc{Klotski} using the open-source MoE models: Mixtral-8$\times$7B and Mixtral-8$\times$22B. They have 46.7B and 141B parameters in bfloat16 precision respectively. We use Mixtral-8$\times$7B and Mixtral-8$\times$22B in environment 1 and use \newtext{Mixtral-8$\times$22B} in environment 2 only. This is because Environment 2 is not considered a resource-constrained environment for Mixtral-8$\times$7B. The inputs are randomly sampled from wikitext-103~\cite{merity2016pointer}, which has rich text from various fields. We use batch sizes from 4 to 64, with a sequence input length of 512 and an output sequence length of 32. We use throughput (generated tokens/generation time) as the metric, where generation time is the total time spent in the prefill and decode phases. We mainly evaluate the throughput of \textsc{Klotski} for different sizes of inputs and compare it with the baselines. The experimental results shown are the average results from multiple trials.

\begin{figure*}
  \centering
  \includegraphics[width=\linewidth]{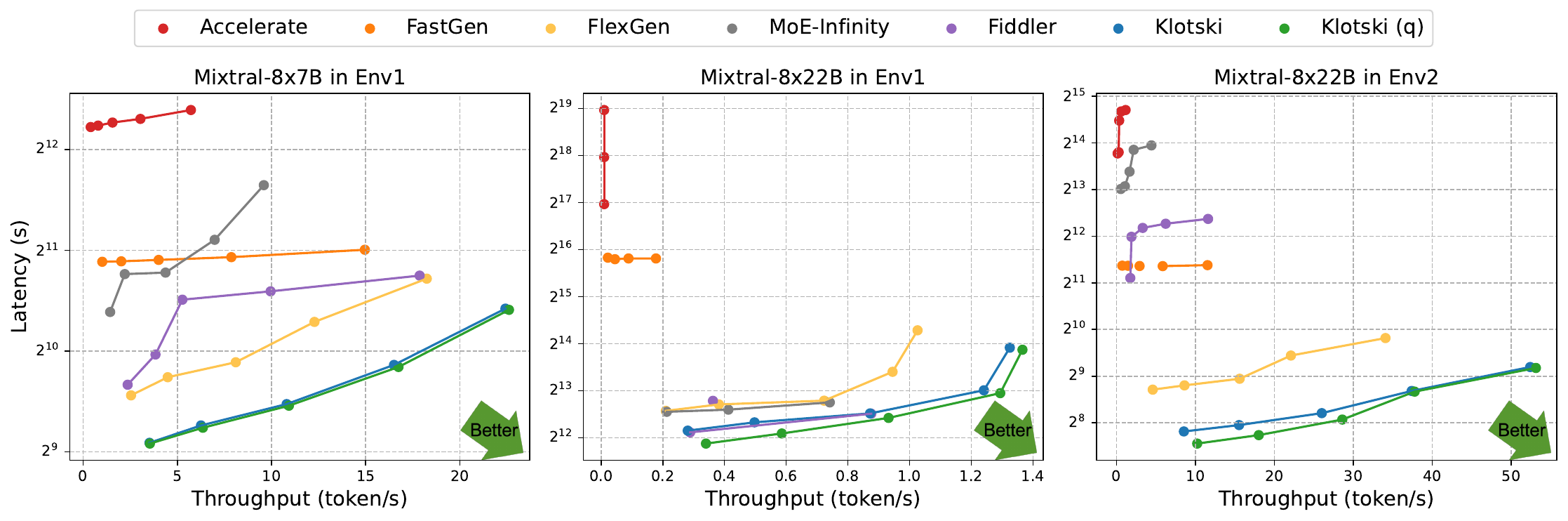}
  \caption{\newtext{Throughput latency trade-off comparison between \textsc{Klotski} and baselines. The curve closer to the lower right is better. (q) means that quantization and dequantization are used.}}
  \label{figure:trade-off}
\end{figure*}

\textbf{Baselines.} 
\newtext{We use the following five offloading studies as baselines for comparison experiments. Among them, the first three works are designed for the dense model, and the last two works are designed for the MoE model.}
\begin{itemize}
\item {\verb|Hugging Face Accelerate|~\cite{accelerate}}: Accelerate supports offloading weights of some layers based on the device map. It's easy to use as a library on Hugging Face Transformers. Hereinafter referred to as Accelerate.
\item {\verb|DeepSpeed-FastGen|~\cite{holmes2024deepspeed}}: It is a version of DeepSpeed ZeRO-Inference after many updates. Since ZeRO-Inference does not support MoE inference, we use DeepSpeed-FastGen which supports Mixtral models. Hereinafter referred to as FastGen.
\item {\verb|FlexGen|~\cite{sheng2023flexgen}}: FlexGen is an efficient offloading work for inference of LLM. It's the first to propose that traverse the computational graph column-by-column.
\item \newtext{{\texttt{Fiddler}~\cite{kamahori2024fiddler}}: In addition to utilizing CPU memory resources, Fiddler uses CPU computing power for MoE model inference, minimizing data movement between the CPU and GPU.}
\item \newtext{{\texttt{MoE-Infinity}~\cite{xue2024moe}}: MoE-Infinity reduces the latency overhead associated with offloading experts through activation-aware expert prefetching and caching.}
\end{itemize}
\newtext{
Additionally, FlexGen only supports dense models with the same structure as OPT, while the others natively support Mixtral. We adapt FlexGen to the Mixtral series of MoE models without changing its primary strategies.}

\subsection{End-to-End Throughput}
\label{section:9.2}
We first evaluate the end-to-end throughput of Klotski and compare it with the baselines, as shown in \autoref{figure:9}. 
\newtext{
We use the maximum $n$ (= 15) from \autoref{figure:impact_of_n} to show a better result than the default computed $n$. And we use $n = 10$ for Mixtral-8$\times$22B in Environment 1 because the computed $n$ is large, which causes out-of-memory (OOM). We set FlexGen to use the same $n$ as us.
Across various scenarios, \textsc{Klotski} consistently outperforms other methods in enhancing MoE inference throughput. 
Compared to Accelerate, FastGen, FlexGen, MoE-Infinity, and Fiddler, \textsc{Klotski} improves the inference throughput by up to 85.12$\times$, 15.45$\times$, 2.23$\times$, 19.06$\times$, and 9.53$\times$, respectively.} 

On Mixtral-8$\times$7B, as the batch size increases, the time difference between computation and I/O gradually narrows, allowing Accelerate and FastGen to achieve decent performance. However, when applied to the larger Mixtral-8$\times$22B, the significantly increased weight transfer leads to a larger time difference between computation and I/O. This ultimately results in throughput that is far inferior to FlexGen and \textsc{Klotski}.

Although FlexGen considers multiple batches and maximizes the use of GPU and CPU memory through tensor slicing, it prefetches the entire MoE layer, requiring a large \( n \) to fully overlap computation and I/O. In contrast, \textsc{Klotski}'s approach to expert prefetching is more flexible, not only compressing inter-layer bubbles but also avoiding additional I/O. Moreover, \textsc{Klotski} further compresses intra-layer bubbles by rearranging the order of expert computations. Additionally, \textsc{Klotski} considers maximizing both memory utilization and transmission speed. \newtext{Furthermore, even if we increase the batch size to 128, Klotski can still achieve a 15\%~($\frac{53 - 46}{46}$) throughput improvement over FlexGen.}

\newtext{
On the other hand, both Fiddler and MoE-Infinity achieve high throughput in Environment 1. Specifically, Fiddler determines that, in Environment 1, performing certain computations on the CPU can be faster than loading and executing them on the GPU. And MoE-Infinity, through its effective prefetching strategy, minimizes unnecessary I/O operations, further optimizing performance. In contrast, \textsc{Klotski} attain higher throughput by effectively overlapping substantial I/O through multiple computations. This underscores that I/O is a critical factor influencing inference latency in offloading-based inference systems.
Moreover, when running inference in Environment 2, both systems show reduced performance as the increased GPU memory and faster I/O make their advantages diminished. In contrast, \textsc{Klotski} orchestrates multi-batch computations, thereby utilizing the GPU more efficiently.
Additionally, when performing Mixtral-8$\times$22B inference on a single 3090, Fiddler and MoE-Infinity are limited to a maximum batch size of 16. This limitation arises because Fiddler and MoE-Infinity only support the offloading of experts. Consequently, the extensive KV cache may result in OOM errors when the batch is large. While \textsc{Klotski} supports more parts of the model to be offloaded, making it more widely applicable.
}

\subsection{Throughput-Latency Trade-off}
We plotted \autoref{figure:trade-off} based on the throughput-related experimental results. It demonstrates that \textsc{Klotski} offers a better throughput-latency trade-off for completing the same workload. Under the same time budget constraint, \textsc{Klotski} can achieve more than three times the throughput of FlexGen (right plot, where latency equals \(2^9\)) and \newtext{outperforms Accelerate, FastGen, MoE-Infinity, and Fiddler}.

Furthermore, as observed in both \autoref{figure:9} and \autoref{figure:trade-off}, quantization has a minimal impact on maximum throughput. However, it enables a more optimal throughput-latency trade-off curve. This improvement is due to the reduced data required for transfer between heterogeneous memory after tensor quantization, resulting in shorter I/O times. Consequently, a smaller \(n\) can achieve full overlap between computation and I/O. This also prevents the dramatic increase in KV cache size as \(n\) grows, especially when the batch size is large.

\begin{figure}
  \centering
  \includegraphics[width=\linewidth]{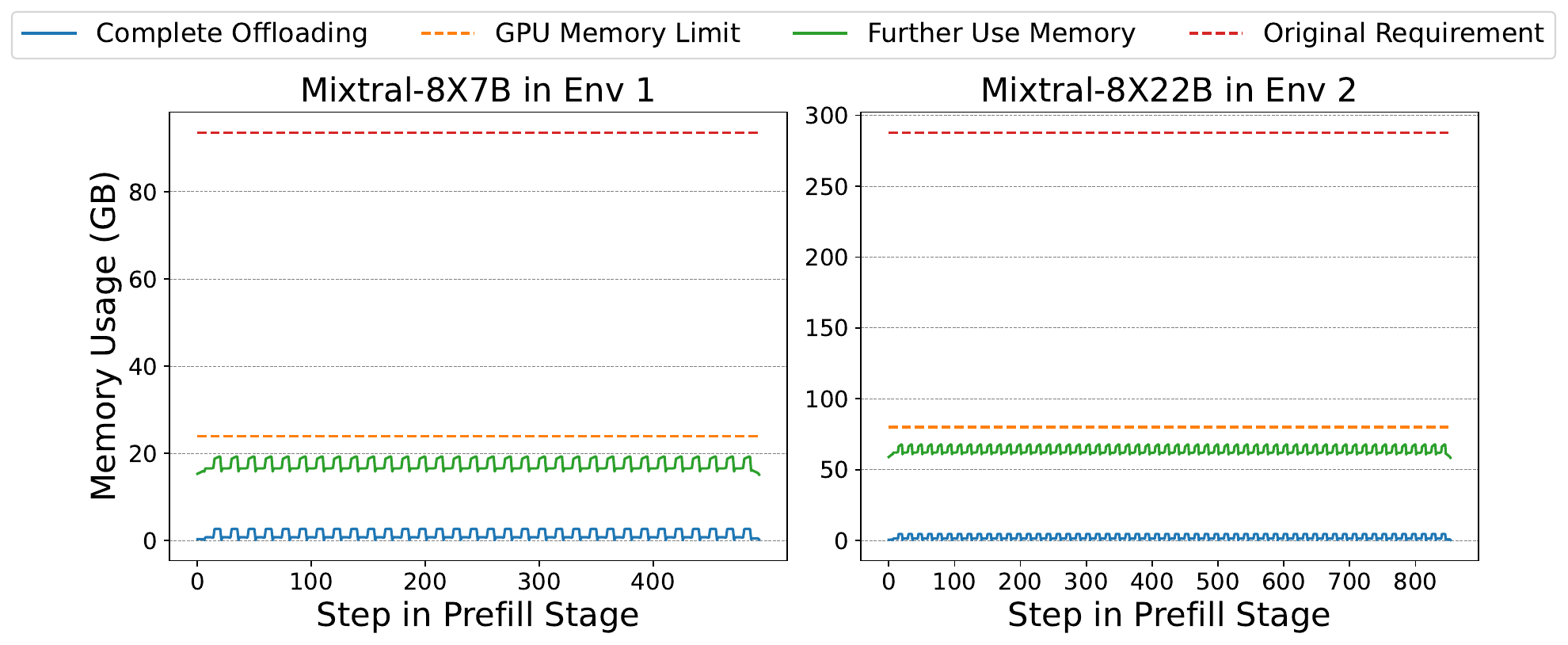}
  \caption{\newtext{GPU memory usage over the prefill. Each step represents one computation of a layer or an expert, i.e. a block in the computation graph as shown in~\autoref{figure:6}.}}
  \label{figure:memory_usage}
\end{figure}

\subsection{Memory Usage}
In \autoref{figure:memory_usage}, we illustrate the GPU memory usage of \textsc{Klotski} during the prefill. \newtext{The red line represents the minimum memory required for inference,} while the \newtext{orange} line indicates the current GPU memory limitation. The blue line shows the memory usage after offloading all tensors, demonstrating that \textsc{Klotski} requires minimal GPU memory to perform MoE inference, reducing memory usage by over 94.1\%. However, there is still a significant amount of expensive GPU memory left unused. Thus, we can further utilize these memory resources, as shown by the \newtext{green} line, achieving a memory reduction of 74.5\% while maintaining a throughput of approximately 40 tokens/s for Mixtral-8$\times$22B on a single H800. The changes during the decoding phase are essentially a repetition of the prefill phase, so for clarity, we only depict the prefill phase in the figure.

\subsection{Ablation Study}
We use that prefetching the entire MoE layer while computing the current layer in the single batch pipeline as a simple pipeline. Building upon this, we achieve our methodology in three steps, comparing the throughput improvements at each step, as shown in \autoref{table:6}. Clearly, considering multi-batch computations provides the most significant enhancement, as it shares weights across multiple batches, significantly reducing inter-layer bubbles. 
\newtext{At the same time, adjusting the computation order of experts leads to reducing intra-layer bubbles, for two main reasons. First, \textsc{Klotski} transfers only hot experts and gate-activated experts, thus avoiding unnecessary expert transfers. Second, this adjustment prioritises the computation of highly requested experts, thereby allowing more time for cold expert transfers, which takes advantage of the high computational demand of hot experts.}
Quantization does not significantly impact the maximum throughput, as its primary function is to reduce I/O overhead, enabling the throughput curve to plateau quickly and reducing the need for larger $n$ values.

\begin{table}
\renewcommand{\arraystretch}{1.4}
\centering
\caption{\newtext{Ablation study of \textsc{Klotski}. The data in the table are throughput (token/s).}}
\resizebox{\linewidth}{!}{
\begin{tabular}{cccc}
\Xhline{1pt}
\multirow{2}{*}{\textbf{Model}}     & \multicolumn{2}{c}{\textbf{Environment 1}}       & \textbf{Environment 2} \\ \cmidrule(lr){2-3} \cmidrule(lr){4-4}
                                    & \textbf{Mixtral-8$\times$7B}   & \textbf{Mixtral-8$\times$22B} & \textbf{Mixtral-8$\times$22B} \\ \hline
\textbf{Simple Pipeline}          & 5.721                   & 0.01                      & 1.149                  \\ \hline
\textbf{+ Multi batches}               & 18.24                   & 0.97                      & 34.07                  \\ \hline
\textbf{+ Only prefetch hot experts}               & 19.074                  & 1.127                       & 44.17 \\ \hline
\textbf{\textsc{Klotski} (+ adjust order)}                      & 22.414                  & 1.325                      & 52.85                  \\ \hline
\textbf{\textsc{Klotski} (q)}                   & 22.604                 & 1.366                     & 53.125                \\ \Xhline{1pt}
\end{tabular}
}
\label{table:6}
\end{table}

\subsection{\newtext{Prefetch Accuracy}}

\begin{figure}
  \centering
  \includegraphics[width=\linewidth]{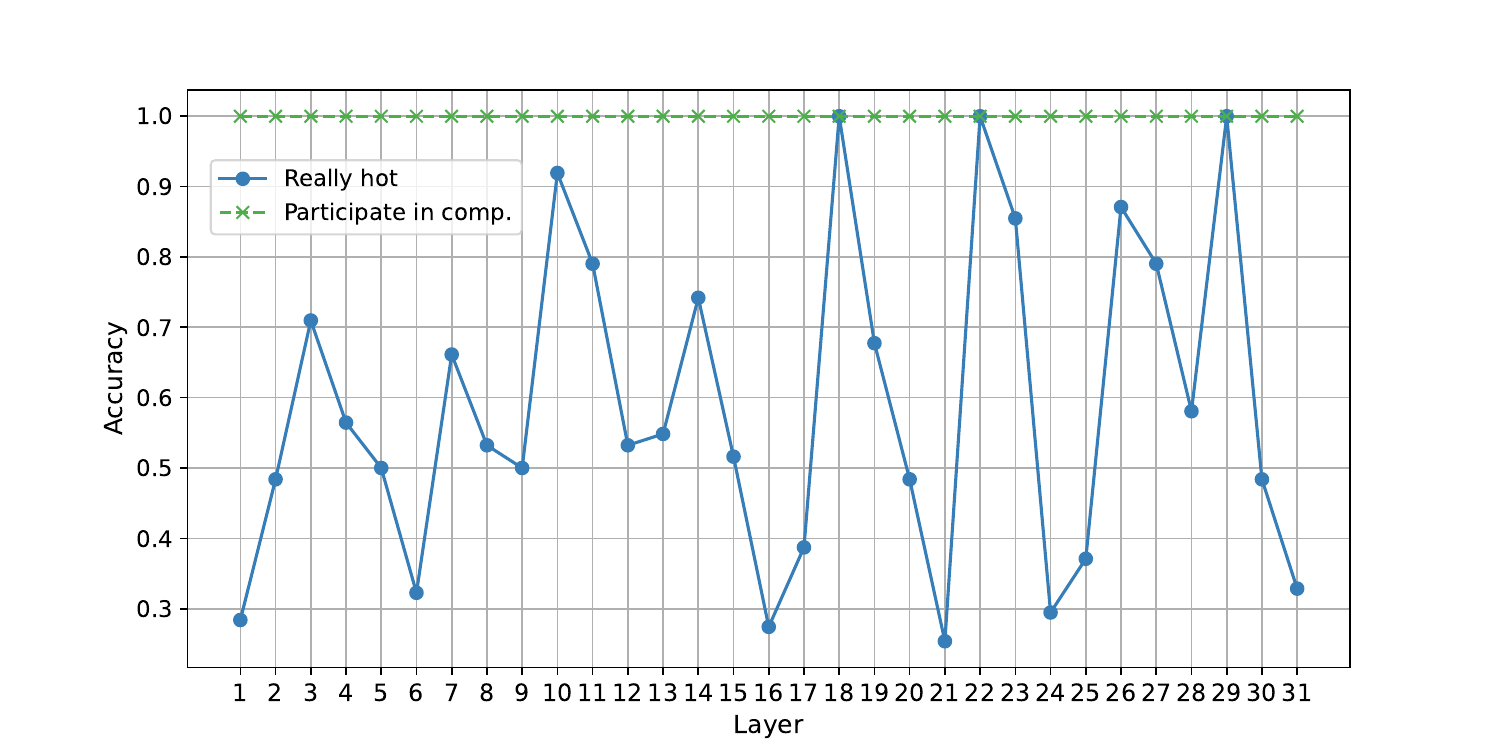}
  \caption{\newtext{Accuracy of prefetched experts per layer. The green line shows how many prefetched hot experts participated in the computation. The blue line indicates the accuracy that the prefetched hot experts are indeed the hot experts of the layer.}}
  \label{figure:accuracy}
\end{figure}

\newtext{
To evaluate the effectiveness of the correlation-aware expert prefetcher, we calculated the accuracy of the prefetched hot experts at each layer, as shown in~\autoref{figure:accuracy}. The green line shows the percentage of prefetched hot experts at each layer that were actually involved in the computations. This result remained consistently at 100\%, demonstrating that \textsc{Klotski} does not transfer experts who are not involved in the computations, thus avoiding unnecessary I/O. In contrast, we also evaluated the average accuracy of prefetching experts for a single sequence, which was found to be 42.24\%. This comparison shows that processing multiple batches simultaneously can effectively reduce I/O waste. In addition, the blue line represents the accuracy of the selected hot experts, which varies with the data, giving an average accuracy of 58.89\%. This suggests that we can accurately predict hot experts in most cases. Furthermore, one of \textsc{Klotski}'s advantages is that \textsc{Klotski} does not rely solely on the accuracy of expert prefetching to overlap I/O. Specifically, \textsc{Klotski} takes a more fine-grained approach to overlap computation and I/O between experts, ensuring that even if a prediction is incorrect, it won't have a big impact.}

\subsection{\newtext{Impacts of $n$ and Batch Size}}

\begin{figure}
  \centering
  \includegraphics[width=\linewidth]{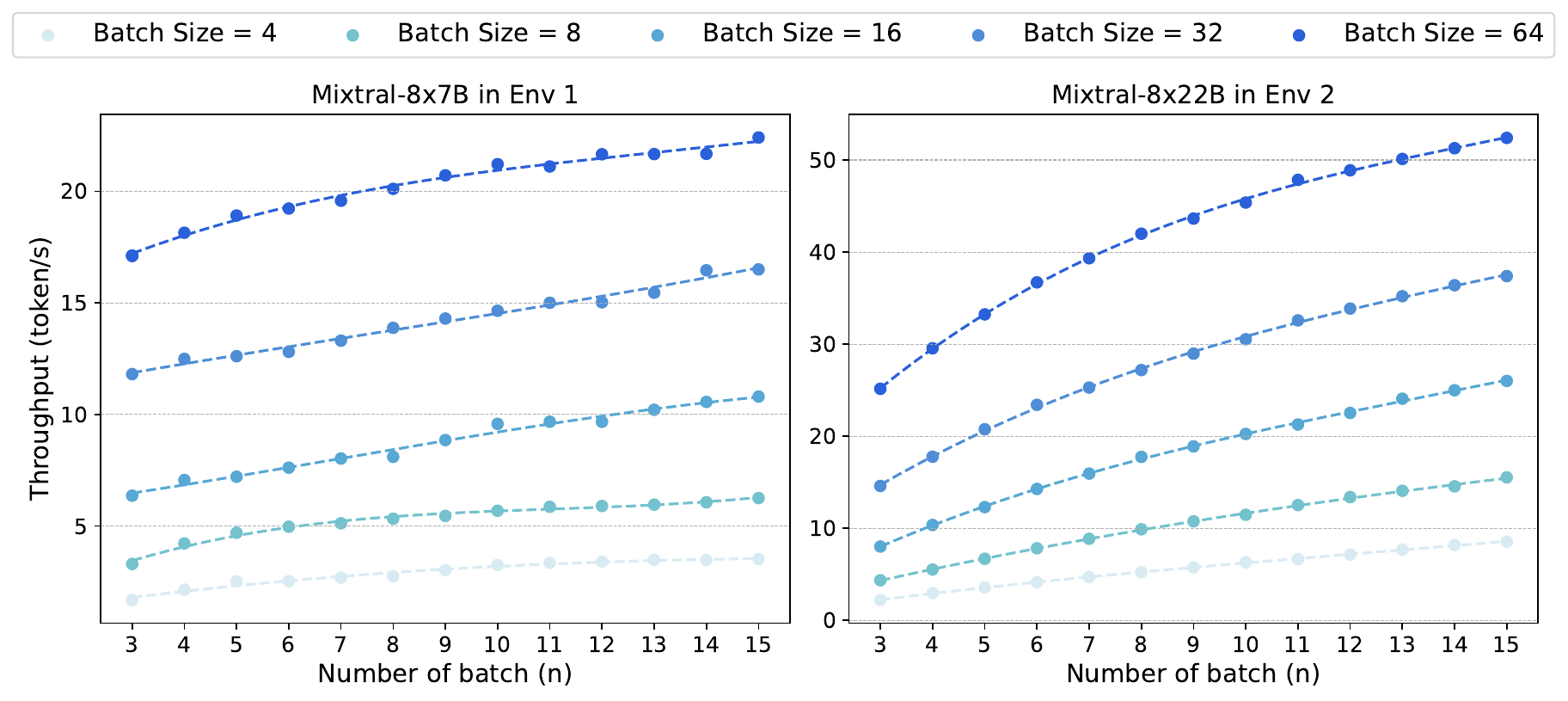}
  \caption{\newtext{The impacts of $n$ and batch size on throughput.}}
  \label{figure:impact_of_n}
\end{figure}

\newtext{
We present detailed end-to-end throughput data, as shown in~\autoref{figure:impact_of_n}, to simulate different scenarios and analyze the impacts of $n$ and batch size on throughput. Due to the large number of GPU hours required to complete all \( n \times bs \) combinations using Mixtral-8$\times$22B in Environment 1, we have not included it here.}

\newtext{
From~\autoref{figure:impact_of_n}, we observed that when \( n \) is small, the throughput is low because the I/O time is much longer than the computation time, resulting in a long latency. In this stage, increasing \( n \) primarily achieves more overlap between computation and I/O. As \( n \) increases, larger batch sizes lead to a faster increase in throughput because each additional batch brings multiple sequences into the computations, further facilitating the overlap of computation and I/O. When \( n \) reaches a sufficiently large value, the slope of the corresponding point on the curve gradually approaches zero, indicating that most of the inter- and intra-layer bubbles have been eliminated. At this stage, increasing \( n \) mainly serves to share the weights among more batches to reduce the number of I/O operations further.}

\subsection{\newtext{Bubble Reduction}}

\begin{figure}
  \centering
  \includegraphics[width=\linewidth]{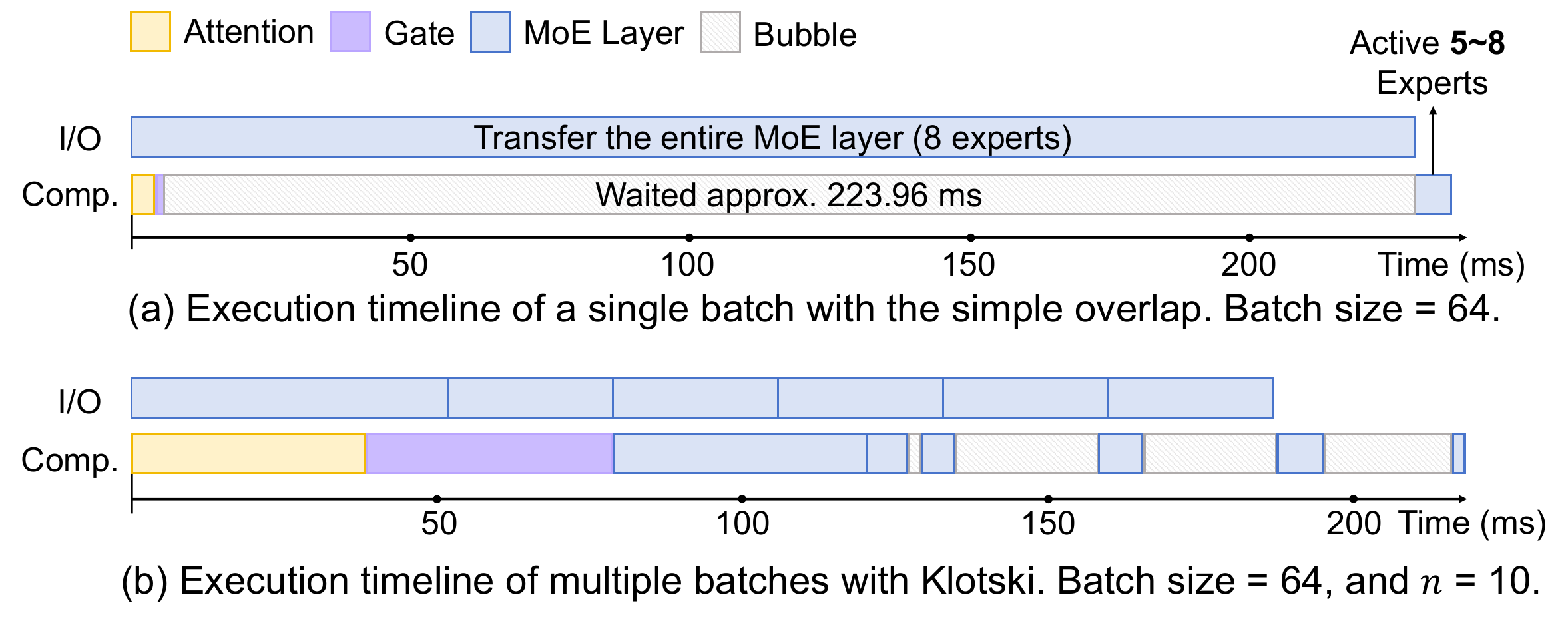}
  \caption{\newtext{Compare the actual pipelines of different methods for performing inference in Environment 1 with Mixtral-8$\times$7B. Simple overlap means prefetching the next layer when executing the current layer. Comp. means computation.}}
  \label{figure:bubbles}
\end{figure}

\newtext{As shown in~\autoref{figure:bubbles}, we proportionally make a detailed inference pipeline of an MoE block, based on the data obtained from using the $profiler$ tool.
\autoref{figure:bubbles}(a) presents the inference pipeline of a single batch using methods designed for dense models. These methods load the entire MoE layer, resulting in significant inter-layer bubbles. In addition, the number of active experts often falls below eight, resulting in unnecessary I/O overhead for loading inactive experts.
In contrast, as shown in \autoref{figure:bubbles}(b), \textsc{Klotski} eliminates inter-layer gaps between the attention and MoE layers. After the gate computation is finished, the hot expert computation starts immediately without waiting. However, due to the significant gap between computation and I/O, it remains challenging to eliminate intra-expert bubbles, even at $n = 10$. By orchestrating expert computations, \textsc{Klotski} overlaps the computation and the I/O between experts, effectively reducing latency. For an identical workload (batch size = 64, number of batches = 10), \textsc{Klotski} completes inference in about 215 ms, compared to approximately 2367 ms using a simple overlap method. 
}

\newtext{
By further increasing $n$, \textsc{Klotski} can achieve the elimination of intra-expert bubbles within the MoE layer,
such as $n = 15$ in~\autoref{section:9.2}.
However, the massively growing KV cache introduces additional load costs, resulting in new bubbles within multi-batch attention layer computations. We aim to address this in future work by developing a generalized and efficient sparse KV cache strategy for \textsc{Klotski}, which will further improve efficiency and achieve a bubble-free multi-batch inference pipeline.}

\section{Conclusion}
We present \textsc{Klotski}, an inference engine designed for MoE models that can perform high-throughput inference in resource-constrained environments. Leveraging the proposed expert-aware multi-batch pipeline paradigm, \newtext{\textsc{Klotski} can significantly reduce the bubbles in the inference pipeline.} Extensive experiments demonstrate that \textsc{Klotski} offers a superior throughput-latency trade-off. For instance, running Mixtral-8$\times$22B inference on a single NVIDIA 3090 achieves a throughput of over 1.3 token/s. Across all experimental scenarios, \textsc{Klotski}'s throughput can be up to 85.12$\times$ greater than that of the existing state-of-the-art.


\bibliographystyle{ACM-Reference-Format}
\balance
\bibliography{sample-base}

\end{document}